\documentclass[conference]{IEEEtran}
\IEEEoverridecommandlockouts
\usepackage{cite}
\usepackage{amsmath,amssymb,amsfonts}
\usepackage{algorithmic}
\usepackage{graphicx}
\usepackage{textcomp}
\makeatletter

\newcommand{\Rmnum}[1]{\expandafter\@slowromancap\romannumeral #1@}
\makeatother
\usepackage{multirow, makecell}
\def\BigRoman{\uppercase\expandafter{\romannumeral\number\count 255 }}
\def\Romannumeral{\afterassignment\BigRoman\count255=}
\usepackage{tikz}
\def\BibTeX{{\rm B\kern-.05em{\sc i\kern-.025em b}\kern-.08em
    T\kern-.1667em\lower.7ex\hbox{E}\kern-.125emX}}

\makeatletter
\def\set@curr@file#1{%
  \begingroup
    \escapechar\m@ne
    \xdef\@curr@file{\expandafter\string\csname #1\endcsname}%
  \endgroup
}
\def\quote@name#1{"\quote@@name#1\@gobble""}
\def\quote@@name#1"{#1\quote@@name}
\def\unquote@name#1{\quote@@name#1\@gobble"}
\makeatother

\begin{document}

\title{Towards Brain-Computer Interfaces for Drone Swarm Control
\footnote{{\thanks{\hrule Research was funded by Institute of Information \& Communications Technology Planning \& Evaluation (IITP) grant funded by the Korea government (No. 2017-0-00451, Development of BCI based Brain and Cognitive Computing Technology for Recognizing User’s Intentions using Deep Learning).}
}}
}

\author{\IEEEauthorblockN{Ji-Hoon Jeong$^1$, Dae-Hyeok Lee$^1$, Hyung-Ju Ahn$^1$, and Seong-Whan Lee$^2$}
\IEEEauthorblockA{$^1$Department of Brain and Cognitive Engineering, Korea University, Seoul, Republic of Korea}
\IEEEauthorblockA{$^2$Department of Artificial Intelligence, Korea University, Seoul, Republic of Korea}

{jh$\_$jeong@korea.ac.kr, lee$\_$dh@korea.ac.kr, hj\_ahn@korea.ac.kr, sw.lee@korea.ac.kr}
}

%\author{\IEEEauthorblockN{}
%\IEEEauthorblockA{{$^1$Department of Brain and Cognitive Engineering, Korea University, Seoul, Republic of Korea} \\
%{$^2$Department of Artificial Intelligence, Korea University, Seoul, Republic of Korea}\\
%}
%}

\maketitle

\begin{abstract}
Noninvasive brain-computer interface (BCI) decodes brain signals to understand user intention. Recent advances have been developed for the BCI-based drone control system as the demand for drone control increases. Especially, drone swarm control based on brain signals could provide various industries such as military service or industry disaster. This paper presents a prototype of a brain-swarm interface system for a variety of scenarios using a visual imagery paradigm. We designed the experimental environment that could acquire brain signals under a drone swarm control simulator environment. Through the system, we collected the electroencephalogram (EEG) signals with respect to four different scenarios. Seven subjects participated in our experiment and evaluated classification performances using the basic machine learning algorithm. The grand average classification accuracy is higher than the chance level accuracy. Hence, we could confirm the feasibility of the drone swarm control system based 
on EEG signals for performing high-level tasks.
\end{abstract}

\begin{small}
\textbf{\textit{Keywords-brain-computer interface; electroencephalogram; drone swarm control; visual imagery}}\\
\end{small}

\section{Introduction}
Brain-computer interface (BCI) analyzes brain signals to understand intention and status of human that can be used for controlling various machines. Since brain signals contain significant information about status of human, many BCI studies have attempted to understand brain signals \cite{C3,MRCP,B1,ECoG2}. In contrast, invasive methods such as electrocroticogram (ECoG) \cite{ECoG} place the electrodes on the brain directly to acquire high-quality brain signals. These methods can obtain the higher quality of brain signals compared with non-invasive methods such as electroencephalogram (EEG), functional near-infrared spectroscopy (fNIRS), and functional magnetic resonance imaging (fMRI), but they are riskier because they involve surgery to implant electrodes. EEG-based BCI has several paradigms for signal acquisition such as motor imagery (MI) \cite{A2,C2,kam}, event-related potential (ERP) \cite{EEG,A1}, and movement-related cortical potential (MRCP) \cite{jeong2020decoding,MRCP}. As applications of EEG-based BCI, a robotic arm \cite{C1,roboticarm,A2}, a speller \cite{speller,won2017motion,ECoG2,stawicki2017novel}, a wheelchair \cite{wheelchair}, and a drone \cite{wang2018wearable,lafleur2013quadcopter,karavas2015effect,karavas2017hybrid} were commonly used for communication between human and machines. 

%%%%%%%%%%%%%%%%%%%%%%%%%%%%%%%%%%%%%%%%%%%%%%%%%%%%%%%%%%%%%%%%%%%%%%
\begin{figure}[t]
\centerline{\includegraphics[width = \columnwidth]{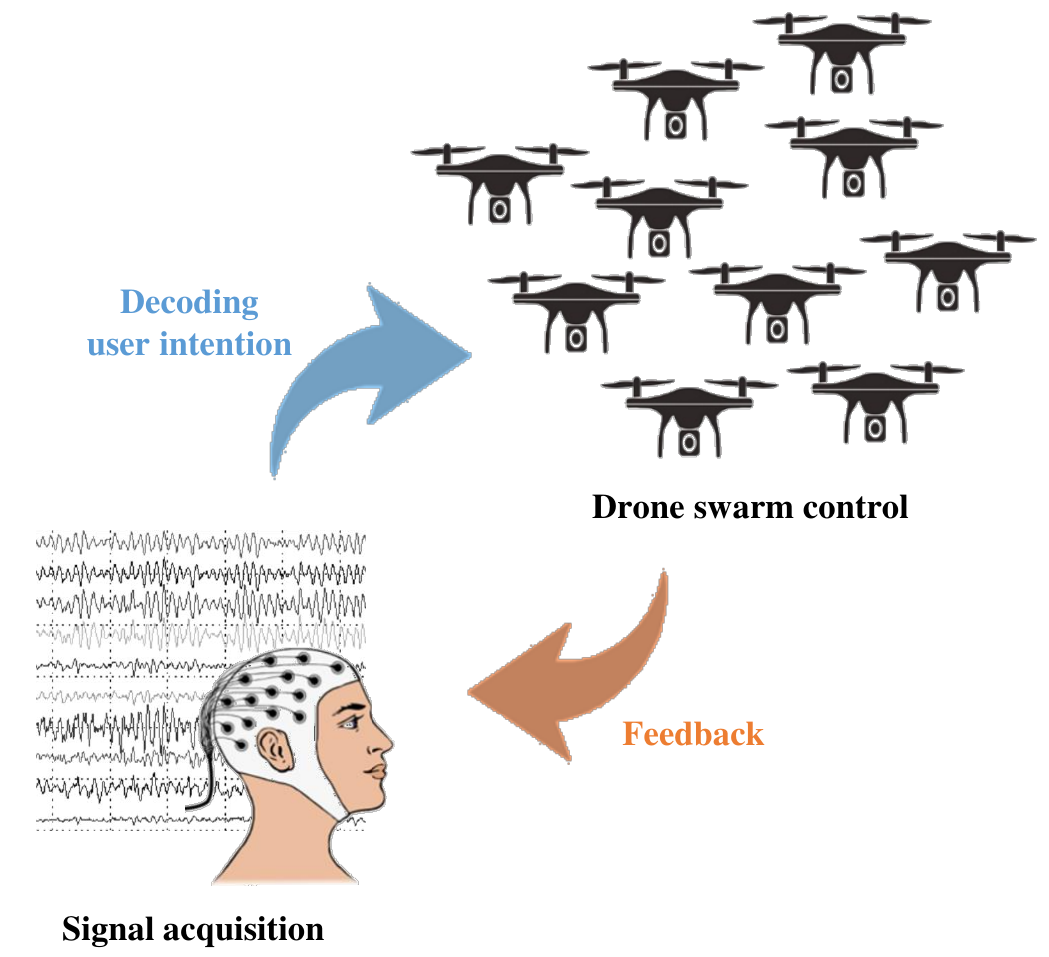}}
\caption{The example of a brain-swarm interface for drone control}
\end{figure}
%%%%%%%%%%%%%%%%%%%%%%%%%%%%%%%%%%%%%%%%%%%%%%%%%%%%%%%%%%%%%%%%%%%%%

Recently, a BCI-based drone swarm is one of the most interesting topics (Fig. 1). Drone swarm means a group of three or more drones. Before drone swarm studies became active, a BCI-based single-unit drone studies had begun first. Wang et al. \cite{wang2018wearable} designed a wearable BCI system based on the steady-state visual evoked potential (SSVEP), which enables three-dimensional navigation of quadcopter flight with visual feedback using a head-mounted device. They demonstrated the feasibility of using the head-mounted device and a proper control strategy to facilitate the portability and practicability of the SSVEP-based BCI system for its navigation utility. LaFleur et al. \cite{lafleur2013quadcopter} reported a novel experiment of BCI controlling a quadcopter in three-dimensional physical space using noninvasive scalp EEG signals in human subjects. They showed the ability to control a flying robot in three-dimensional physical space with EEG signals. Very few research groups have studied to control a drone swarm. Karavas et al. \cite{karavas2015effect} examined the perception and representation of collective behaviors of swarms at the brain level of human supervisors. They extracted event related potentials at EEG signals. Their study provided the first evidence of representation of swarm collective behaviors at the brain level which can lead to the design of a new generation of brain-swarm control and perception interfaces. Karavas et al. \cite{karavas2017hybrid} proposed a hybrid BCI system which combined EEG signals and joystick input. The purpose of applying this system was to show both the system’s capability for control of actual robotic platforms and the feasibility of controlling robotic swarm behaviors using EEG signals. They used event related desynchronization / synchronization phenomena. Their study allowed for continuous control variables extracted from the EEG signals.

In this study, we measured EEG signals of 4-class using visual imagery paradigm \cite{sousa2017pure,koizumi2018development}: `Hovering', `Splitting', `Dispersing', and `Aggregating'. The classes used in the experimental paradigm consist of the most basic commands for controlling a drone swarm. To best of our knowledge, this is the first attempt that demonstrates the feasibility of classifying the high-level commands which consist of 4-class. Second, we achieved robust classification performance in the 4-class high-level commands compared with the chance-level accuracy (0.25).

The rest of this paper is organized as follows. Section {\Romannumeral 2} gives a description of the experimental protocols, EEG signals acquisition, a drone swarm control simulator and the data analysis. Section {\Romannumeral 3} presents the results of performance accuracies for 4-class classification and discussions about our study. In session {\Romannumeral 4}, conclusion and future works are described.\\

\section {Materials and Methods}
\subsection{Experimental Protocols}
Seven healthy subjects, who were naive  BCI users, have recruited in the experiment (aged 22-33, five males and two females). Before the experiment, each subject was informed of the experimental protocols and procedures. After they had understood, all of them provided their given written consent according to the Declaration of Helsinki. All experimental protocols and environments were reviewed and approved by the Institutional Review Board at Korea University (KUIRB-2020-0013-01). First, the subjects sat in front of the experimental desk as depicted in Fig. 2. The monitor display for visual instruction was put at the distance of 90cm from the subjects. The subjects were asked to perform visual imagery according to the four different scenarios. Each trial was composed of four phases such as rest, visual cue/preparation, stare fixation point, and imagination. In the rest phase, the subject took a comfortable rest with restraining eye and body movement for 3 s. After the rest phase, the monitor displayed one of the scenarios as a visual cue and then subjects prepared the visual imagery task according to the cue. Then, the subjects stared the fixation point during 3 s to avoid an afterimage effect. During the 4 s, The subjects conducted visual imagery task. We asked subjects to perform 200 trials in total (i.e., 50 trials 4 classes) (Fig. 3).

%%%%%%%%%%%%%%%%%%%%%%%%%%%%%%%%%%%%%%%%%%%%%%%%%%%%%%%%%%%%%%%%%%%%%%
\begin{figure}[t]
\centerline{\includegraphics[scale=0.8]{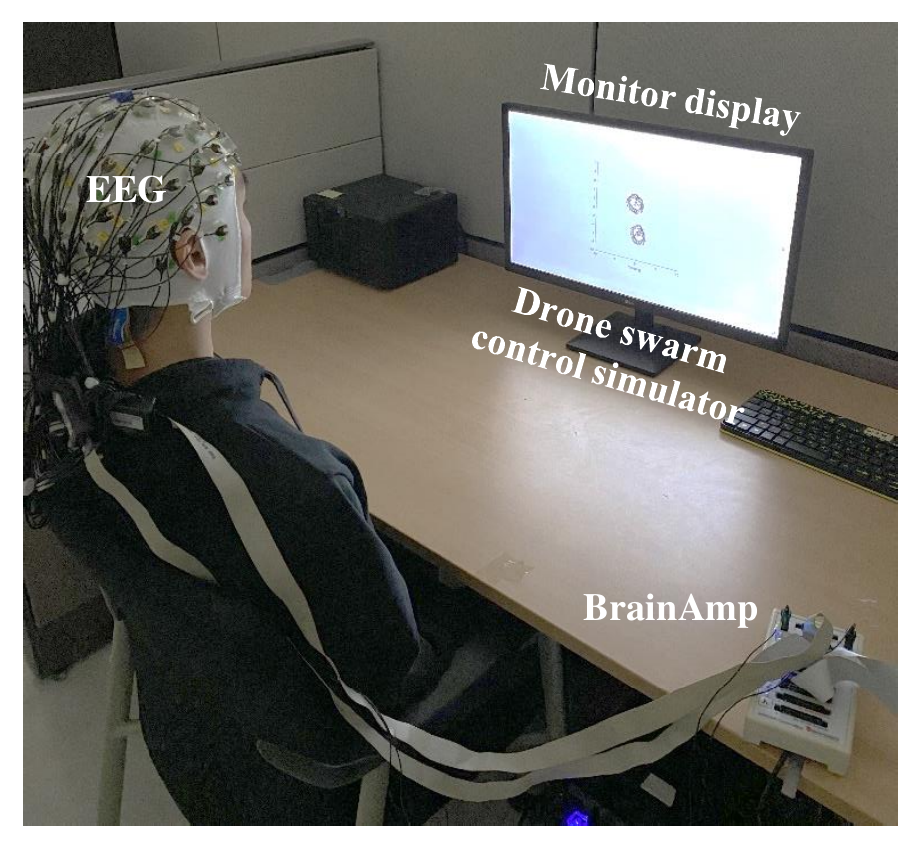}}
\caption{Experimental environment.}
\end{figure}
%%%%%%%%%%%%%%%%%%%%%%%%%%%%%%%%%%%%%%%%%%%%%%%%%%%%%%%%%%%%%%%%%%%%%
%%%%%%%%%%%%%%%%%%%%%%%%%%%%%%%%%%%%%%%%%%%%%%%%%%%%%%%%%%%%%%%%%%%%%%
\begin{figure}[t]
\centerline{\includegraphics[width = \columnwidth]{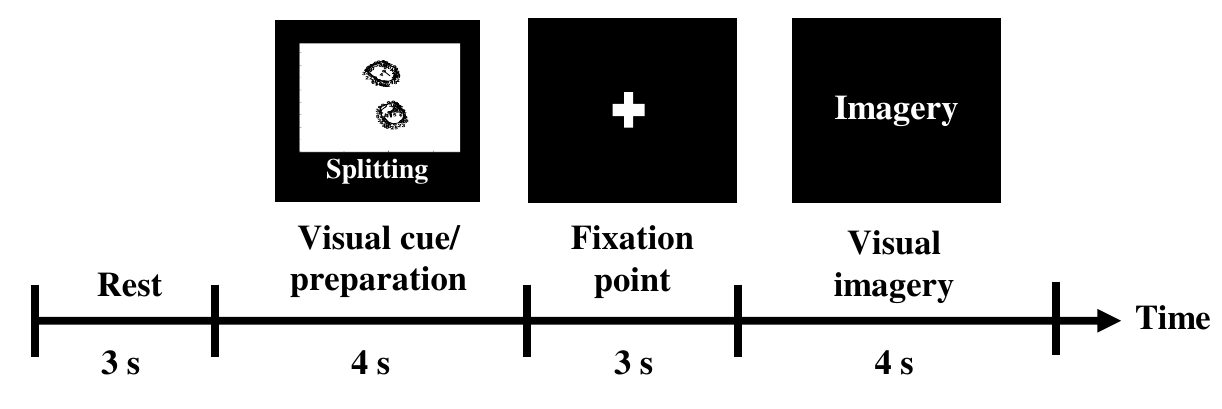}}
\caption{Experimental paradigm for visual imagery.}
\end{figure}
%%%%%%%%%%%%%%%%%%%%%%%%%%%%%%%%%%%%%%%%%%%%%%%%%%%%%%%%%%%%%%%%%%%%%

\subsection{EEG Signal Acquisition}
We acquired the EEG signals with respect to drone swarm control scenarios using BrainVision Recorder (BrainProducts GmbH, Germany). EEG signals were acquired using 64 Ag/AgCl electrodes following 10/20 international systems. The ground and reference channels were FCz and FPz positions, respectively. The sampling rate was 1,000 Hz, and a notch filter was applied to the acquired signals as 60 Hz. All electrode impedances were kept below 10 k$\Omega$ during the experiment (Fig. 4). 

%%%%%%%%%%%%%%%%%%%%%%%%%%%%%%%%%%%%%%%%%%%%%%%%%%%%%%%%
\begin{figure*}[t]
\centerline{\includegraphics[scale=0.9]{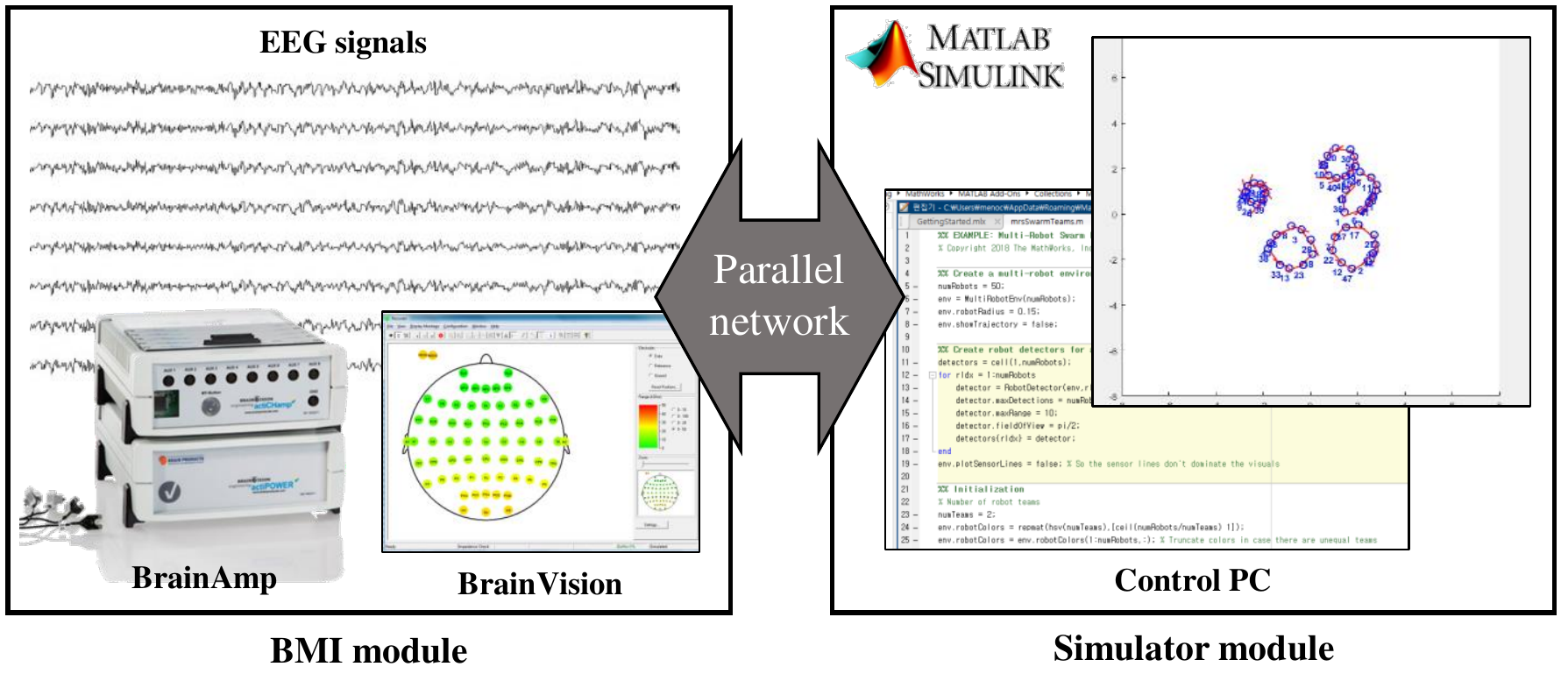}}
\caption{System architecture for EEG data acquisition with a drone swarm simulator environment.}
\end{figure*}
%%%%%%%%%%%%%%%%%%%%%%%%%%%%%%%%%%%%%%%%%%%%%%%%%%%%%%%%%%%%%%
%%%%%%%%%%%%%%%%%%%%%%%%%%%%%%%%%%%%%%%%%%%%%%%%%%%%%%%%%%%%%%
\begin{figure*}[t]
\centerline{\includegraphics[width = \textwidth]{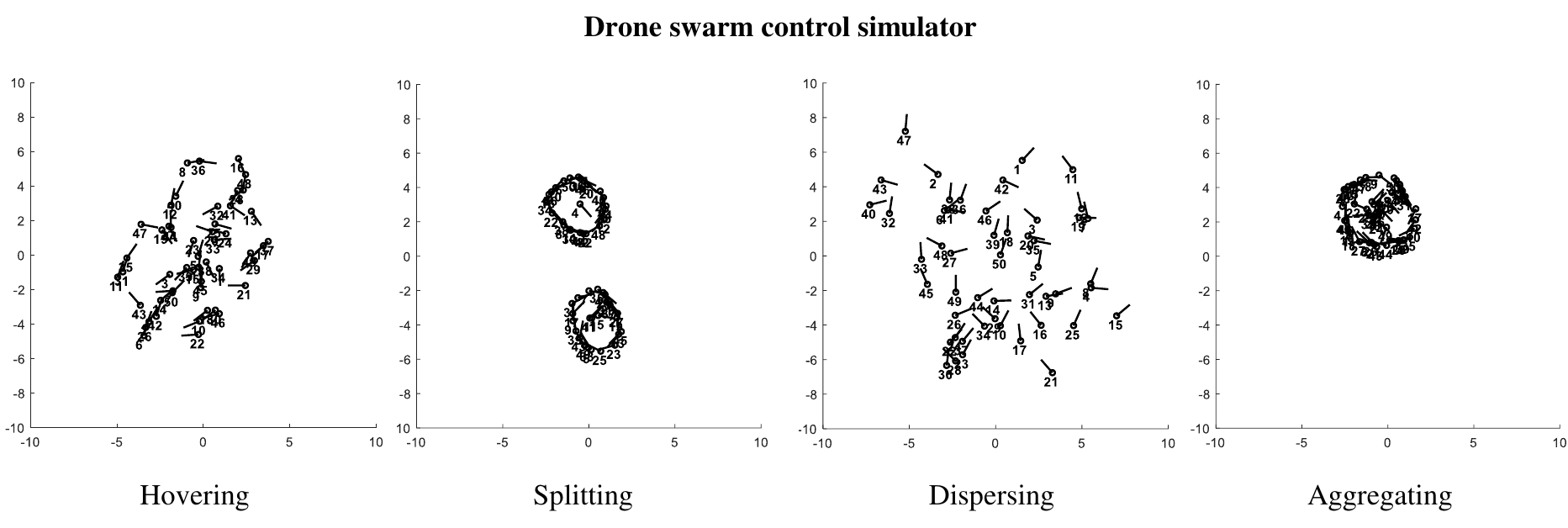}}
\caption{The example representation of a drone swarm simulator for four different scenarios.}
\end{figure*}
%%%%%%%%%%%%%%%%%%%%%%%%%%%%%%%%%%%%%%%%%%%%%%%%%%%%%%%%%%%%%%

\subsection{Drone Swarm Control Simulator}
We designed a drone swarm control simulator using Matlab software (MathWorks, USA) with Mobile Robotics Simulation Toolbox. This toolbox provides utilities for robot simulation and algorithm development in the 2D grid maps. We modified a multi-robot lidar control to drone swarm control system which was composed of fifty unit drones as depicted in Fig. 3. The drone conducted four different scenarios such as `Hovering', `Splitting', `Dispersing', and `Aggregating' through the simulator (Fig. 5). The hovering cue indicated an initial position of the swarm drone. The subjects imagined the visual instruction for the hovering state of drones. The splitting cue showed that the swarm drone divided two different swarms. The dispersing cue represented randomly position of each unit drone with outspreading. Finally, the aggregating cue indicated the unit drones was positioned closely with each other. One of the scenarios showed to the subjects during the `Visual cue/preparation' phase.

\subsection{Data Analysis}
For the acquired EEG data verification, we adopted a basic EEG classification procedure which is generally used conventional BCI studies \cite{C3,MRCP,B1,ECoG2}. The data were preprocessed by using a band-pass filter with a zero-phase 2nd Butterworth filter between [8-30] Hz. The spectral ranges also mostly used in the imagination decoding from EEG signals which is mu and beta band. We segmented the data into 4 s epoched data for each trial. Then, a common spatial pattern (CSP) algorithm \cite{FBCSP} was applied to extract dominant spatial features for training. A transformation matrix from CSP consisted of the logarithmic variances of the first three and the last three columns were used as a feature. A linear discriminant analysis (LDA) \cite{channel} was used for a classification method which classified four different class using one-versus-rest strategy. For a fair evaluation of classification performance, a 5-fold cross-validation was used.

\section {Results and discussion}

As Fig. 6, the grand-average classification accuracy for four different scenarios is 36.7 ($\pm$4.6)\% across all subjects. 
Subject 5 showed the highest classification performance as 41.3\%, but subject 3 indicated the lowest results as 28.4\%. However, those accuracies are higher than the chance level accuracy for the 4-class classification problem (approximately 25\%). That means, although we used the basic machine learning algorithm for evaluating classification performances, we could confirm that the EEG data acquired with high quality under a restrained environment. Actually, the subject 3, who showed the lowest performance, tended to have difficulty with the visual imagery task during the experiment. Through the experiment, we confirm that visual instructions in a form similar to the real-world environment are necessary such as actual drone swarm control. It could be more helpful the subjects could perform visual imagery tasks.

%%%%%%%%%%%%%%%%%%%%%%%%%%%%%%%%%%%%%%%%%%%%%%%%%%%%%%%%
\begin{figure}[t]
\centerline{\includegraphics[width = \columnwidth]{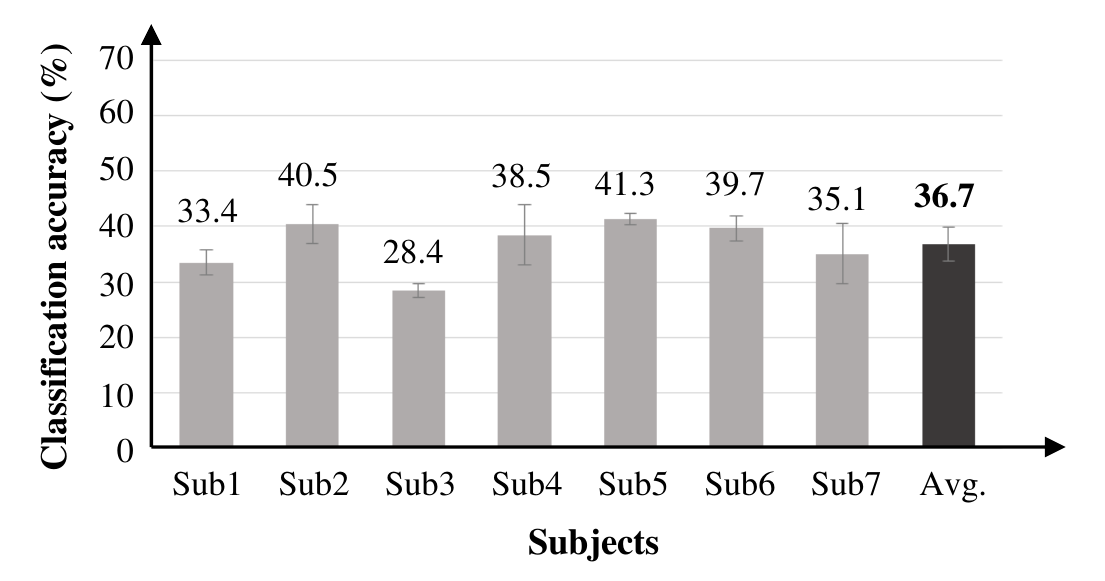}}
\caption{Classification accuracies of the four scenarios across all subjects.}
\end{figure}
%%%%%%%%%%%%%%%%%%%%%%%%%%%%%%%%%%%%%%%%%%%%%%%%%%%%%%%%%%%%%%

\section{Conclusion and Future works}
In this paper, we designed an experimental environment for acquiring EEG data with respect to visual imagery tasks. Through the experiment system, the subjects could perform the visual imagery for the drone swarm control of various scenarios. We have implemented the four main different classes for swarm flight such as `Hovering', `Splitting', `Dispersing', and `Aggregating'. These scenarios have been used as an important function of drone swarm control under a simulated wargame environment. The EEG classification performance of visual imagery achieved a little higher than the chance rate level yet, this experiment system could contribute to developing a brain-swarm interface system using the drone for military service, industrial disaster, and artificial intelligence development.

Hence, we will have investigated a drone swarm control system based on EEG signals it could possible to conduct high-level tasks. As a result, the EEG classification performance needs to be higher and be cover more multi-command. Therefore, we will adopt the deep learning approach to our developing system for drone swarm control robustly under real-world environments. It would greatly improve the interaction effect between the user and the drone.

\section{Acknowledgement}
The authors thanks to B.-H. Kwon for their help with the design of experimental paradigm.\\
\bibliographystyle{IEEEbib}
\bibliography{BCI-Drone}

\begin{thebibliography}{10}

\bibitem{C3}
C.~I. Penaloza and S.~Nishio,
\newblock ``B{MI} control of a third arm for multitasking,''
\newblock {\em Sci. Robot.}, vol. 3, no. 20, pp. eaat1228, 2018.

\bibitem{MRCP}
I.~K. Niazi, N.~Jiang, O.~Tiberghien, J.~F. Nielsen, K.~Dremstrup, and
  D.~Farina,
\newblock ``Detection of movement intention from single-trial movement-related
  cortical potentials,''
\newblock {\em J. Neural Eng.}, vol. 8, no. 6, pp. 066009, 2011.

\bibitem{B1}
{X. Ding and S.-W. Lee},
\newblock ``Changes of functional and effective connectivity in smoking
  replenishment on deprived heavy smokers: a resting-state f{MRI} study,''
\newblock {\em PLoS One}, vol. 8, no. 3, pp. e59331, 2013.

\bibitem{ECoG2}
M.-H. Lee, J.~Williamson, D.-O. Won, S.~Fazli, and S.-W. Lee,
\newblock ``A high performance spelling system based on {EEG}-{EOG} signals
  with visual feedback,''
\newblock {\em IEEE Trans. Neural Syst. Rehabil. Eng.}, vol. 26, no. 7, pp.
  1443--1459, 2018.

\bibitem{ECoG}
{G. Buzs{\'a}ki, C. A. Anastassiou, and C. Koch},
\newblock ``The origin of extracellular fields and currents—{EEG}, {EC}o{G},
  {LFP} and spikes,''
\newblock {\em Nat. Rev. Neurosci.}, vol. 13, no. 6, pp. 407, 2012.

\bibitem{A2}
J.-H. Kim, F.~Bie{\ss}mann, and S.-W. Lee,
\newblock ``Decoding three-dimensional trajectory of executed and imagined arm
  movements from electroencephalogram signals,''
\newblock {\em IEEE. Trans. Neural. Syst. Rehabil. Eng.}, vol. 23, no. 5, pp.
  867--876, 2014.

\bibitem{C2}
J.-H. Jeong, K.-T. Kim, D.-J. Kim, and S.-W. Lee,
\newblock ``Decoding of multi-directional reaching movements for {EEG}-based
  robot arm control,''
\newblock in {\em IEEE Int. Conf. Syst., Man, and Cybern. (SMC)}, 2019, pp.
  511--514.

\bibitem{kam}
T.-E. Kam, H.-I. Suk, and S.-W. Lee,
\newblock ``Non-homogeneous spatial filter optimization for
  electroencephalogram {(EEG)}-based motor imagery classification,''
\newblock {\em Neurocomputing}, vol. 108, pp. 58--68, 2013.

\bibitem{EEG}
Y.~Chen, A.~D. Atnafu, I.~Schlattner, W.~T. Weldtsadik, M.-C. Roh, H.~J. Kim,
  S.-W. Lee, B.~Blankertz, and S.~Fazli,
\newblock ``A high-security {EEG}-based login system with {RSVP} stimuli and
  dry electrodes,''
\newblock {\em IEEE Inf. Fore. Sec.}, vol. 11, no. 12, pp. 2635--2647, 2016.

\bibitem{A1}
X.~Zhu, H.-I. Suk, S.-W. Lee, and D.~Shen,
\newblock ``Canonical feature selection for joint regression and multi-class
  identification in {A}lzheimer’s disease diagnosis,''
\newblock {\em Brain imaging behav.}, vol. 10, no. 3, pp. 818--828, 2016.

\bibitem{jeong2020decoding}
J.-H. Jeong, N.-S. Kwak, C.~Guan, and S.-W. Lee,
\newblock ``Decoding movement-related cortical potentials based on
  subject-dependent and section-wise spectral filtering,''
\newblock {\em IEEE Trans. Neural Syst. Rehabil. Eng.}, 2020.

\bibitem{C1}
J.~Meng, S.~Zhang, A.~Bekyo, J.~Olsoe, B.~Baxter, and B.~He,
\newblock ``Noninvasive electroencephalogram based control of a robotic arm for
  reach and grasp tasks,''
\newblock {\em Sci. Rep.}, vol. 6, pp. 38565, 2016.

\bibitem{roboticarm}
J.-H. Jeong, K.-H. Shim, J.-H. Cho, and S.-W. Lee,
\newblock ``Trajectory decoding of arm reaching movement imageries for
  brain–controlled robot arm system,''
\newblock in {\em Annual Int. Conf. IEEE Eng. Med. and Bio. Soc. (EMBC)}, 2019,
  pp. 23--27.

\bibitem{speller}
D.-O. Won, H.-J. Hwang, S.~D{\"a}hne, K.~R. M{\"u}ller, and S.-W. Lee,
\newblock ``Effect of higher frequency on the classification of steady-state
  visual evoked potentials,''
\newblock {\em J. Neural Eng.}, vol. 13, no. 1, pp. 016014, 2015.

\bibitem{won2017motion}
D.-O. Won, H.-J. Hwang, D.-M. Kim, K.-R. M{\"u}ller, and S.-W. Lee,
\newblock ``Motion-based rapid serial visual presentation for gaze-independent
  brain-computer interfaces,''
\newblock {\em IEEE Trans. Neural Syst. Rehabil. Eng.}, vol. 26, no. 2, pp.
  334--343, 2017.

\bibitem{stawicki2017novel}
P.~Stawicki, F.~Gembler, A.~Rezeika, and I.~Volosyak,
\newblock ``A novel hybrid mental spelling application based on eye tracking
  and {SSVEP}-based {BCI},''
\newblock {\em Brain Sci.}, vol. 7, no. 4, pp. 35, 2017.

\bibitem{wheelchair}
K.-T. Kim, H.-I. Suk, and S.-W. Lee,
\newblock ``Commanding a brain-controlled wheelchair using steady-state
  somatosensory evoked potentials,''
\newblock {\em IEEE. Trans. Neural. Syst. Rehabil. Eng.}, vol. 26, no. 3, pp.
  654--665, 2016.

\bibitem{wang2018wearable}
M.~Wang, R.~Li, R.~Zhang, G.~Li, and D.~Zhang,
\newblock ``A wearable {SSVEP}-based {BCI} system for quadcopter control using
  head-mounted device,''
\newblock {\em IEEE Access}, vol. 6, pp. 26789--26798, 2018.

\bibitem{lafleur2013quadcopter}
K.~LaFleur, K.~Cassady, A.~Doud, K.~Shades, E.~Rogin, and B.~He,
\newblock ``Quadcopter control in three-dimensional space using a noninvasive
  motor imagery-based brain-computer interface,''
\newblock {\em J. Neural Eng.}, vol. 10, no. 4, pp. 046003, 2013.

\bibitem{karavas2015effect}
G.~K. Karavas and P.~Artemiadis,
\newblock ``On the effect of swarm collective behavior on human perception:
  {T}owards brain-swarm interfaces,''
\newblock in {\em IEEE IEEE Int. Conf. Multisen. Fusion Integ. Intell. Syst.
  (MFI)}, 2015, pp. 172--177.

\bibitem{karavas2017hybrid}
G.~K. Karavas, D.~T. Larsson, and P.~Artemiadis,
\newblock ``A hybrid {BMI} for control of robotic swarms: {P}reliminary
  results,''
\newblock in {\em IEEE/RSJ Int. Conf. Intell. Robot. and Syst. (IROS)}, 2017,
  pp. 5065--5075.

\bibitem{sousa2017pure}
T.~Sousa, C.~Amaral, J.~Andrade, G.~Pires, U.~J. Nunes, and M.~Castelo-Branco,
\newblock ``Pure visual imagery as a potential approach to achieve three
  classes of control for implementation of {BCI} in non-motor disorders,''
\newblock {\em J. Neural Eng.}, vol. 14, no. 4, pp. 046026, 2017.

\bibitem{koizumi2018development}
K.~Koizumi, K.~Ueda, and M.~Nakao,
\newblock ``Development of a cognitive brain-machine interface based on a
  visual imagery method,''
\newblock in {\em Annual Int. Conf. IEEE Eng. Med. and Bio. Soc. (EMBC)}, 2018,
  pp. 1062--1065.

\bibitem{FBCSP}
K.~K. Ang, Z.~Y. Chin, H.~Zhang, and C.~Guan,
\newblock ``Filter bank common spatial pattern ({FBCSP}) in brain-computer
  interface,''
\newblock in {\em IEEE Int. Joint Conf. on Neural Netw.}, 2008, pp. 2390--2397.

\bibitem{channel}
J.-H. Cho, J.-H. Jeong, K.-H. Shim, D.-J. Kim, and S.-W. Lee,
\newblock ``Classification of hand motions within {EEG} signals for
  non-invasive {BCI}-based robot hand control,''
\newblock in {\em IEEE Int. Conf. Syst., Man, and Cybern. (SMC)}, 2018, pp.
  515--518.

\end{thebibliography}

\end{document}